\documentclass[letterpaper, 10 pt, conference]{ieeeconf}  

\IEEEoverridecommandlockouts                              

\usepackage{amsmath} 
\usepackage{amssymb}  
\usepackage{arydshln}
\usepackage{etoc}

\usepackage{float}
\usepackage{graphicx}
\usepackage{booktabs}
\usepackage{multirow}
\usepackage{multicol}
\usepackage{booktabs}
\usepackage{multirow}
\usepackage{multicol}
\usepackage{bm}
\usepackage{color, colortbl}
\usepackage{subcaption}
\usepackage[dvipsnames,x11names]{xcolor}
\usepackage{etoc}
\usepackage{float}

\DeclareMathOperator*{\argmin}{arg\,min}
\let\labelindent\relax
\usepackage{enumitem}
\title{\LARGE \bf
SynH2R: Synthesizing Hand-Object Motions for Learning Human-to-Robot Handovers
}

\author{Sammy Christen$^{1*}$, Lan Feng$^{1*}$, Wei Yang$^{2}$, Yu-Wei Chao$^{2}$, Otmar Hilliges$^{1}$, Jie Song$^{1}$  
\thanks{$^{1}$Department of Computer Science,
        ETH Zurich $^{2}$NVIDIA%
}
}

\begin{document}

\maketitle
\thispagestyle{empty}
\pagestyle{empty}

\newcommand{\StatesSet}{\mathcal{S}}
\newcommand{\ActionsSet}{\mathcal{A}}
\newcommand{\TransitionSet}{\mathcal{T}}
\newcommand{\GoalsSet}{\mathcal{G}}
\newcommand{\state}[1]{{s}_{#1}}
\newcommand{\goal}[1]{{g}_{#1}}
\newcommand{\action}[1]{{a}_{#1}}
\newcommand{\reward}{r}
\newcommand{\discountRate}{\gamma}
\newcommand{\QFunction}{Q}
\newcommand{\Cost}{J}
\newcommand{\mdp}{\mathcal{M}}
\newcommand{\taskname}{dynamic grasp synthesis}
\newcommand{\methodname}{\textit{D-Grasp}\xspace}
\newcommand{\supmat}{supp.~material\xspace}
\newcommand{\raisim}{RaiSim\xspace}
\newcommand{\CIRCLE}[1]{\raisebox{.5pt}{\footnotesize\textbf{\textcircled{\raisebox{-.6pt}{\textnormal{\textcolor{black}{#1}}}}}}}
\newcommand{\gc}{\cellcolor[HTML]{E4E4E4}}



\newcommand{\figref}[1]{Fig.~\ref{#1}}
\newcommand{\secref}[1]{Section~\ref{#1}}
\newcommand{\algref}[1]{Algorithm~\ref{#1}}
\newcommand{\eqnref}[1]{Eq.~\eqref{#1}}
\newcommand{\tabref}[1]{Tab.~\ref{#1}}
\newcommand{\update}[1]{{\color{black}{#1}}}
\newcommand{\camera}[1]{{\color{black}{#1}}}
\newcommand{\Fig}[1]{Fig.~\ref{fig:#1}}
\newcommand{\Figure}[1]{Figure~\ref{fig:#1}}
\newcommand{\Tab}[1]{Tab.~\ref{tab:#1}}
\newcommand{\Table}[1]{Table~\ref{tab:#1}}
\newcommand{\eq}[1]{(\ref{eq:#1})}
\newcommand{\Eq}[1]{Eq.~\ref{eq:#1}}
\newcommand{\Equation}[1]{Equation~\ref{eq:#1}}
\newcommand{\Sec}[1]{Sec.~\ref{sec:#1}}
\newcommand{\Section}[1]{Section~\ref{sec:#1}}
\newcommand{\Appendix}[1]{Appendix~\ref{app:#1}}
\newcommand{\etal}{et al.}

\DeclareRobustCommand*\circledorange[1]{\tikz[baseline=(char.base)]{
            \node[hexorange,shape=circle,draw,inner sep=1pt, thick=8pt, scale=0.8] (char) {#1};}}
       
\DeclareRobustCommand*\circledblue[1]{\tikz[baseline=(char.base)]{
            \node[hexblue,shape=circle,draw,inner sep=1pt, thick=8pt, scale=0.8] (char) {#1};}}

\newif\ifshowcomments
\showcommentstrue 

\makeatletter
\def\and{
  \end{tabular}%
  \hskip 2.85em \@plus.17fil%
  \begin{tabular}[t]{c}}
\makeatother

\ifshowcomments
        \newcommand{\note}[3]{{\textcolor{#2}{[#1: #3]}}}
        \newcommand{\OH}[1]{\note{Otmar}{green}{#1}}
		\newcommand{\SC}[1]{\note{Sammy}{teal}{#1}}
        \newcommand{\wei}[1]{{\color{blue} [WEI: #1] }}
        \newcommand{\ywchao}[1]{{\color{red} [Yu-Wei: #1] }}
        \newcommand{\claudia}[1]{{\color{magenta} [Claudia: #1] }}
        \newcommand{\dieter}[1]{{\color{olive} [Dieter: #1] }}
        \newcommand{\TL}[1]{{\color{cyan} [TL: #1] }}

        \newcommand{\todo}[1]{\textcolor{red}{#1}}
        \newcommand{\edit}[1]{\textcolor{magenta}{#1}}
\else
        \newcommand{\note}[3]{\unskip}
        \newcommand{\OH}[1]{\unskip}
		\newcommand{\SC}[1]{\unskip}
		\newcommand{\ywchao}[1]{\unskip}
		\newcommand{\wei}[1]{\unskip}
		\newcommand{\claudia}[1]{\unskip}
		\newcommand{\dieter}[1]{\unskip}
        \newcommand{\TL}[1]{{\unskip}}
\fi

\newcommand{\oh}[1]{\OH{#1}}
\newcommand{\jh}[1]{\JH{#1}}

\newcommand{\posevec}{\mathbf{q}_h}
\newcommand{\posevecgen}{\mathbf{q}}
\newcommand{\posevechand}{\mathbf{q_h}}
\newcommand{\posvec}{\mathbf{x}}
\newcommand{\goalvec}{\mathbf{g}}
\newcommand{\vertexvec}{\mathbf{v}}
\newcommand{\posesix}{\mathbf{T}}
\newcommand{\replay}{\mathbf{\mathcal{D}}}
\newcommand{\goals}{\mathbf{G}}
\newcommand{\forcevec}{\mathbf{f}}
\newcommand{\actionvec}{\mathbf{u}}
\newcommand{\actions}{\mathbf{a}}
\newcommand{\expertvec}{\mathbf{e}}
\newcommand{\imagevec}{\mathbf{I}}
\newcommand{\pcvec}{\mathbf{p}}
\newcommand{\rewardvec}{\mathbf{r}}
\newcommand{\statevec}{\mathbf{s}}
\newcommand{\transitionvec}{\mathbf{d}}
\newcommand{\torques}{\boldsymbol{\tau}}
\newcommand{\policy}{\boldsymbol{\pi}}
\newcommand{\grasppred}{\boldsymbol{\sigma}}
\newcommand{\policypre}{\boldsymbol{\pi_{\text{pre}}}}
\newcommand{\criticpre}{\boldsymbol{Q_{\text{pre}}}}
\newcommand{\policyft}{\boldsymbol{\pi_*}}
\newcommand{\criticft}{\boldsymbol{Q_*}}
\newcommand{\policyexp}{\boldsymbol{\pi_{\text{exp}}}}
\newcommand{\gpolicyvec}{\boldsymbol{\pi}_{g}}
\newcommand{\mpolicyvec}{\boldsymbol{\pi}_{m}}

\newcommand{\boldparagraph}[1]{\vspace{0.2cm}\noindent{\bf #1}}


\renewcommand{\thefootnote}{\fnsymbol{footnote}}
\newcommand\blfootnote[1]{%
  \begingroup
  \renewcommand\thefootnote{}\footnote{#1}%
  \addtocounter{footnote}{-1}%
  \endgroup
}

\newcolumntype{C}[1]{>{\centering\let\newline\\\arraybackslash\hspace{0pt}}m{#1}}
\newcolumntype{L}[1]{>{\raggedright\let\newline\\\arraybackslash\hspace{0pt}}m{#1}}
\newcolumntype{R}[1]{>{\raggedleft\let\newline\\\arraybackslash\hspace{0pt}}m{#1}}

\newcommand{\greencheck}{{\color{Green4} \checkmark}}
\newcommand{\redcross}{{\color{red}$\times$}}

\begin{abstract}

Vision-based human-to-robot handover is an important and challenging task in human-robot interaction. Recent work has attempted to train robot policies by interacting with dynamic virtual humans in simulated environments, where the policies can later be transferred to the real world. However, a major bottleneck is the reliance on human motion capture data, which is expensive to acquire and difficult to scale to arbitrary objects and human grasping motions. In this paper, we introduce a framework that can generate plausible human grasping motions suitable for training the robot. To achieve this, we propose a hand-object synthesis method that is designed to generate handover-friendly motions similar to humans. This allows us to generate synthetic training and testing data with 100x more objects than previous work. In our experiments, we show that our method trained purely with synthetic data is competitive with state-of-the-art methods that rely on real human motion data both in simulation and on a real system. In addition, we can perform evaluations on a larger scale compared to prior work. With our newly introduced test set, we show that our model can better scale to a large variety of unseen objects and human motions compared to the baselines. 

\end{abstract}

\def\thefootnote{*}\footnotetext{Indicates equal contribution}
\section{Introduction}

 Humans handing over objects to robots is a crucial task in human-robot interaction (HRI)~\cite{ortenzi:tro2021}. Seamless human-to-robot handovers (H2R) will enable robots to assist humans in many domains, such as manufacturing settings, elderly homes, or rehabilitation. In unknown scenarios, robots will encounter objects and human behavior that they have not previously experienced. Therefore, robots should be flexible in handling unseen objects and human behavior.
 
Collecting training experiences for robots in the real world is prohibitively inefficient and unsafe for humans. Therefore, recent research on H2R handovers~\cite{chao:icra2022,christen2023handover} has trained robot policies in simulation by allowing the robot to interact with a simulated human partner, and later transfer the trained policies onto real-world platforms. While this improves the scalability of collecting training experiences for the robot, the pipeline for simulating the human counterpart remains challenging to scale. In order to simulate realistic human motions for handovers, prior work~\cite{chao:icra2022,christen2023handover,pang:roman2021, christen:handshake2019} relies on motion capture data of hand-object interactions~\cite{chao:cvpr2021}. The simulated environment for training robots is thus bounded by the object instances and human motions pre-captured in the mocap dataset. To train on novel objects or human motions, a tedious re-capturing of data with a mocap setup is required. This begs the question: can we automatically synthesize human handover motions on arbitrary objects for robot handover training, and thereby fully leverage the blessing of scalability from training in simulation?

 \begin{figure}
\begin{center}
   \includegraphics[width=0.40\textwidth]{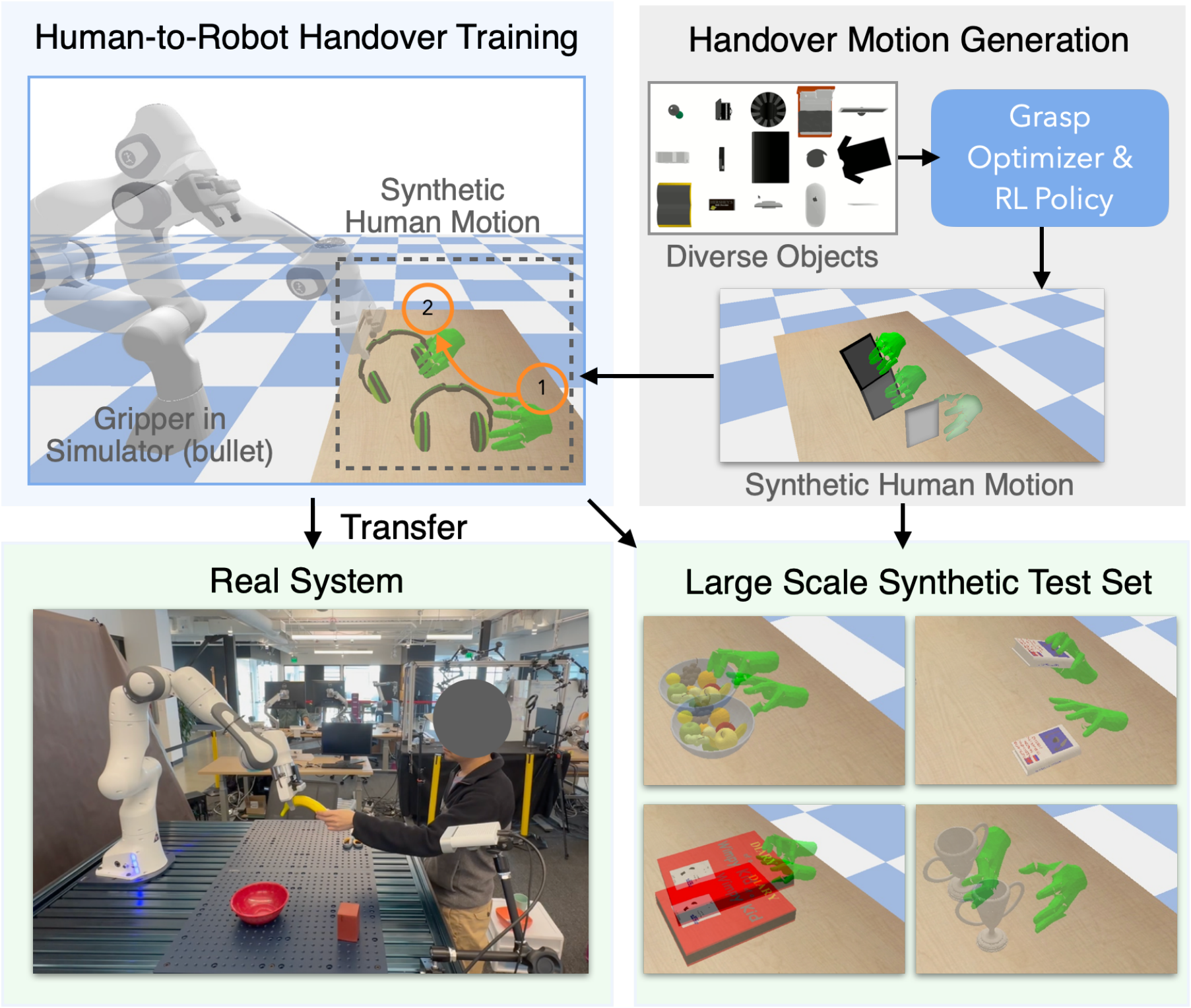}
\end{center}
   \vspace{-0.25cm}
   \caption{Overview of our framework. We train a robot to perform human-to-robot handovers using synthetic human motions. We transfer to a real robot and evaluate on a large synthetic test set of unseen objects and human motions.}
   \vspace{-0.9em}
\label{fig:teaser}
\end{figure}
Fortunately, recent progress in hand-object interaction synthesis \cite{Zhang2021ManipNetNM,christen2022dgrasp} holds the promise to generate natural and physically plausible human grasping motions, which can potentially alleviate the need for expensive motion capture. For example, D-Grasp \cite{christen2022dgrasp} generates hand motions that grasp an object and move it to a target pose using a reinforcement learning (RL) based policy. Despite their promise, these methods are still not readily applicable for human-to-robot handovers. For example, D-Grasp assumes a grasp pose reference as input and does not account for the handover-friendliness of such a grasp. For successful handovers, it is crucial to control the direction of approaching and the amount of free area for the robot to grasp the object. 

In this paper, we combine human-to-robot handover training with hand-object motion synthesis. We build upon D-Grasp~\cite{christen2022dgrasp} and propose a method that can generate natural human grasping motions that are suited for training robots without requiring any high-quality motion capture data. The first question is how to generate grasp references. Current static grasp generation pipelines do not offer controllability with respect to the grasp direction. This makes them unsuitable for handover since humans tend to hand over objects in a direction toward the robot and leave free space on the opposite side for the robot to grasp the object. Besides, we empirically discovered that off-the-shelf learning-based grasp generation models often struggle to generalize to objects beyond the training distribution. This limits their use on arbitrary object datasets without additional training. To this end, we propose an optimization-based grasp generation method that is conditioned on the approaching direction and incentivizes a stable human hand grasp that does not enclose an object fully. Since our grasp generation pipeline is non-learning based, it also does not suffer from generalization issues on unseen objects. We then generate hand pose references on a large set of objects, and pass them to D-Grasp to generate human grasping and handover motions. To improve the grasping of unseen objects, we also augment D-Grasp to condition on an object shape representation. With this pipeline, our method can synthesize diverse human motions for grasping unseen objects at a larger scale. This in turn allows us to leverage much more diverse human motions and objects in simulation to train the robot.
In our experiments, we first evaluate our approach on the HandoverSim benchmark \cite{chao:icra2022}. We demonstrate that training our method from purely synthetic human motion data can achieve on-par performance with recent work that relies on high-quality motion capture data and uses the test objects during training. Furthermore, we introduce a new synthetic test set of 1174 unseen objects which exceeds the scale of previous benchmarks by 100x (see \tabref{tab:related}). Our method outperforms the state-of-the-art baselines on this more challenging testbed. Lastly, we show that users do not recognize any significant differences between a policy trained on purely synthetic data versus a policy trained on real motion capture data, indicating the naturalness and plausibility of our generated human motions. This is an important insight that has implications for the training of robotic agents with simulated humans in the future.

To summarize, our contributions are as follows: i) A new framework to scale up human-to-robot handover training by generating large-scale synthetic human handover motions. ii) A method to generate natural human grasping motions that can scale to many objects and allow control of the direction of approach. iii) Experiments in simulation and on a real system showing our method can perform on par with baselines that use high-quality motion capture data for training. iv) A new synthetic test set that allows the evaluation of human-to-robot handovers on more than 1,000 unknown objects. Our evaluations show that our method outperforms baselines on this new benchmark.
\section{Related Work}
\subsection{Grasp Synthesis for Dexterous Hands}

The problem of dexterous grasp synthesis involves determining optimal grasping poses given an object's mesh or point cloud and is generally categorized into two techniques: \textit{non-learning-based} and \textit{learning-based}. 

\textit{Learning-based} methods employ neural networks to predict grasps, leveraging motion capture grasp datasets~\cite{Jiang2021HandObjectCC,Karunratanakul2020GraspingFL,taheri:eccv2020} or synthetic datasets generated by their non-learning-based counterparts \cite{Miller2004GraspitAV,hasson:cvpr2019}. These methods predominantly rely on conditional variational autoencoder architectures, where the resulting grasps are stochastically sampled from latent space. To generalize to unseen shapes, the objects need to be close enough to the training distribution. In our approach, we use a non-learning based solution and hence do not suffer from generalization issues. \textit{Non-learning-based} methods have been employed to generate extensive synthetic datasets~\cite{Miller2004GraspitAV,Wang2022DexGraspNetAL,turpin2022grasp,turpin2023fast,hasson19obman}. \cite{Miller2004GraspitAV} employs collision detection algorithms to formulate stable grasps. In a different vein, some approaches employ differentiable force closure estimation~\cite{Wang2022DexGraspNetAL, Liu2021SynthesizingDA} for grasp generation. Works such as \cite{turpin2022grasp,turpin2023fast} exploit a differentiable simulation to synthesize grasps. In contrast to these works, our method can be conditioned on the grasp direction and does not rely on any simulator or force closure estimation.

Beyond static grasp synthesis, there are works~\cite{christen2022dgrasp,zhou2022toch,Xu2023UniDexGraspUR,Wu2022LearningGD,braun2023physically, zhang2024artigrasp} that focus on the temporal aspect of hand-object synthesis. D-Grasp \cite{christen2022dgrasp} introduces dynamic grasp synthesis to model hand-object interaction sequences, while \cite{Xu2023UniDexGraspUR} proposes a universal grasp policy generalizable to diverse objects. These methods utilize grasp reference poses to guide their RL-based policies. In our work, we present an RL policy that can be trained on a small set of YCB objects and generalize to unseen objects at inference time. We achieve this by combining grasp references from our non-learning based grasp optimization with an RL-based policy. 

\begin{table}
\centering
\begin{center}
\caption{Comparison between the number of objects and test configurations used in the most related works and ours.}
\vspace{-0.1cm}
\resizebox{0.42\textwidth}{!}{%
\begin{tabular}{l|cc}
   \toprule
    Benchmark & Num Objects & Test Configurations \\
    \midrule
    Sanchez-Matilla \etal~\cite{sanchez-matilla:ral2020} & ~~~4 & ~288 \\ 
    Rosenberger \etal~\cite{rosenberger:ral2021} & ~~13  & ~520 \\
    Yang \etal~\cite{yang:icra2021}    &  ~~26  & ~156 \\
    HandoverSim ('S0') \cite{chao:icra2022}    &  ~~18 & ~144 \\
    \textbf{Ours} &  \textbf{1174} & \textbf{4436} \\
   \bottomrule
   
\end{tabular}
}
\label{tab:related}
\end{center}
\vspace{-0.5cm}

\end{table}
\begin{figure*}[t]
\begin{center}
   \includegraphics[width=0.85\textwidth]{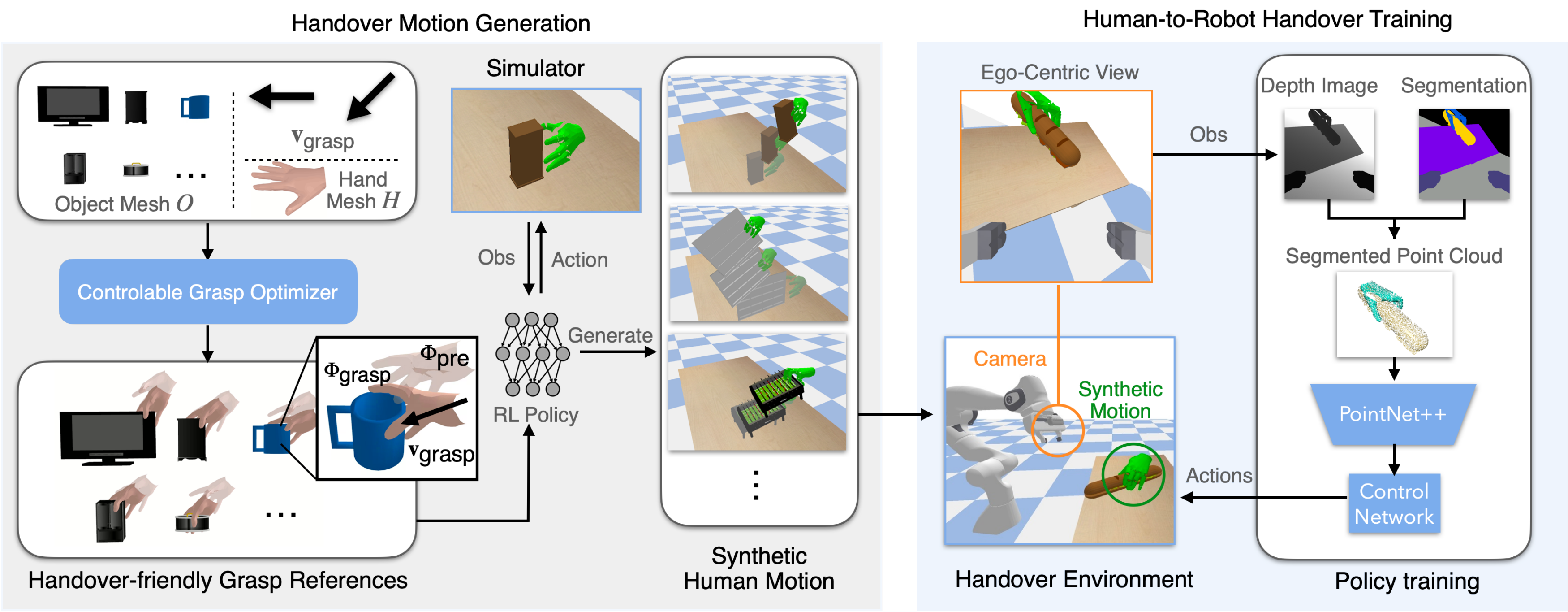}
\end{center}
   \vspace{-0.3cm}
   \caption{\textbf{Method Overview.} Our framework contains a handover motion generation stage and a H2R handover training stage.}
   \vspace{-0.9em}
\label{fig:method_overview}
\end{figure*}


\subsection{Human-to-Robot Handovers}
Recent advances in human-to-robot (H2R) handover systems \cite{chao:icra2022, christen2023handover, pang:roman2021, sanchez-matilla:ral2020, rosenberger:ral2021, yang:icra2021, yang:icra2022, duan:tcds} show the potential of creating robust human-to-robot interaction frameworks. This progress has been driven by the surge of hand-object interaction datasets \cite{chao:cvpr2021, taheri:eccv2020, liu:cvpr2022, garcia-hernando:cvpr2018,hampali:cvpr2020,brahmbhatt:eccv2020,moon:eccv2020,ye:iccv2021,fan2023arctic}, which allows studying H2R handovers as a grasp planning problem \cite{Miller2004GraspitAV, bicchi:icra2000, bohg:tro2014}. These approaches require the exact knowledge of the 3D object shape, and hence do not generalize to unseen objects. To mitigate this, recent works leverage learning-based grasp predictions from vision input ~\cite{rosenberger:ral2021, yang:icra2021,yang:icra2022,duan:tcds, marturi:2019dynamic}. Rosenberger \etal~\cite{rosenberger:ral2021} use hand and object tracking and a grasp selection network to plan H2R handovers, which are executed in open-loop fashion. Yang \etal~\cite{yang:icra2021} propose a reactive H2R system that can generalize to unseen objects by selecting temporally consistent 6DoF grasps from GraspNet \cite{mousavian:iccv2019}. In \cite{yang:icra2022}, this work is improved by employing an MPC-based algorithm that adds reachability criteria to the motion planning. However, these methods either require the human hand to be stationary, complex hand-designed cost functions, and expertise in robot motion planning.  Chao \etal~\cite{chao:icra2022} introduce HandoverSim, a benchmark to evaluate handover policies in simulation. GA-DDPG \cite{wang:corl2021b} propose a vision-based method for grasping static objects, which can be deployed for H2R handovers. However, their method has difficulties in dynamic scenes with humans. Closest to our work, Christen \etal~\cite{christen2023handover} propose a framework to learn vision-based handover policies by training with human grasping motions from the DexYCB dataset \cite{chao:cvpr2021}. In contrast, our work uses synthetic handover motions generated by our method. Therefore, it does not require any real-world mocap data and allows scaling to a much more diverse set of training objects and motions.




\section{Overview}
The goal of this work is to teach a robot agent to perform human-to-robot handovers by training purely on synthetic human motion data. The simulation setting follows HandoverSim \cite{chao:icra2022} and consists of a tabletop scene with different objects, a robot, and a simulated human hand. The robot comprises a 7-DoF Panda arm with a two-fingered gripper and a wrist-mounted RGB-D camera. The simulated hand replays human handover motions (either from motion capture or synthetic), i.e., grasping an object and moving it to a handover location. The goal of the robot is to grasp the object from the human, without collision or dropping, and move it to a designated goal location.
Our framework comprises two stages, as shown in \figref{fig:method_overview}. In the handover motion generation stage (left), we generate synthetic human-object interaction data over a large set of different objects. In the human-to-robot handover training stage, we leverage the synthetic data to train a vision-based human-to-robot handover policy in simulation, which can be transferred to a real system. 

\section{Synthetic Handover Motion Generation}
To synthesize human handover motions (\figref{fig:method_overview} left), we first generate handover-friendly static grasp poses and then utilize these grasps as references to guide an RL-based policy inspired by D-Grasp \cite{christen2022dgrasp} to generate handover motions. 

\subsection{Grasp Reference Generation}
\label{sec:static_grasp_gen}
Our controllable grasp optimizer takes as input an object mesh $\bm{{O}}$, a unified human hand mesh $\bm{{H}}$ given by the parametric MANO hand model \cite{MANO:SIGGRAPHASIA:2017}, and a grasp direction $\mathbf{v}_{\text{grasp}}$ defined by a three-dimensional vector pointing from the wrist joint to the object center. The MANO hand model \cite{MANO:SIGGRAPHASIA:2017} is parameterized by a set of pose parameters $\bm{\theta} \in \mathbb{R}^{45}$, global wrist translation $\bm{\tau} \in \mathbb{R}^{3}$ and global wrist orientation $\bm{\phi} \in \mathbb{R}^{3}$. The pose parameters comprise 15 joints with 3DoF each. We use the lower dimensional Principal Component Analysis (PCA) space with 15 components to represent the pose parameters $\bm{\theta}$. As shown in \figref{fig:method_overview} (left), we introduce a controllable grasp optimizer that outputs two keyframe hand poses—a \textit{pre-grasp pose} $\bm{\Phi}_{\text{pre}} = [\bm{\tau}_{\text{pre}}, \bm{\phi}_{\text{pre}}, \bm{\theta}_{\text{pre}}]$ and a \textit{grasp reference pose} $\bm{\Phi}_{\text{grasp}} = [\bm{\tau}_{\text{grasp}}, \bm{\phi}_{\text{grasp}}, \bm{\theta}_{\text{grasp}}]$.
We first generate the pre-grasp pose, which is crucial to achieving stable grasping \cite{dasari2023icra}, and then optimize the grasp reference pose based on it. Both the pre-grasp pose and the grasp reference pose should adhere to the grasp direction $\mathbf{v}_{\text{grasp}}$. \\

\noindent\textit{1) Pre-Grasp Pose} $\bm{\Phi}_{\text{pre}}.$ We separately optimize the 6D global pre-grasp pose, which includes the global wrist translation $\bm{\tau}_{\text{pre}}$ and wrist orientation $\bm{\phi}_{\text{pre}}$, and the local finger pre-grasp pose $\bm{\theta}_{\text{pre}}$. As shown in \figref{fig:gripper}, we define the line connecting the middle fingertip to the thumb tip as the \textit{grasp axis}. Similarly, the vector originating from the wrist joint and pointing towards the midpoint of this connecting line is termed the hand's \textit{heading}. 
\begin{itemize}[left=0.0em]
\item \textbf{Gripper-like Finger Pose $\bm{\theta}_{\text{pre}}$}. A two-fingered gripper typically has two adjustable fingers designed to grasp an object firmly from both sides. As illustrated in \figref{fig:gripper}, this characteristic is emulated by considering the thumb as one finger of the gripper and grouping the other fingers as the second finger. The gripper-like finger pose $\bm{\theta}_{\text{pre}}$ is derived by maximizing the separation between the thumb tip and the palm plane, i.e., the grasp axis. Specifically, the MANO hand is initialized in the flat hand pose $\bm{\theta}_{\text{flat}}$ without any rotation or translation. By default, the flat hand is positioned in the xz-plane, with the palm oriented in the -y direction. The y-coordinate of the thumb tip, $\mathbf{p}^y_{\text{thumb}} (\bm{\theta})$, is minimized to maximize its separation from the other fingers, resulting in gripper-like finger poses:
\begin{equation}
\bm{\theta}_{\text{pre}} = \mathop{\argmin}\limits_{\bm{\theta}}  
\mathbf{p}^y_{\text{thumb}}(\bm{\theta};\bm{\theta}_{\text{flat}}).
\end{equation}

We use the Adam optimizer~\cite{kingma:2015iclr} with a learning rate set at 0.003 for 300 iterations. Since we optimize in PCA space, the remaining four fingers converge to a natural pose, even though the objective focuses on the thumb's position.

\begin{figure}
\begin{center}
   \includegraphics[width=0.42\textwidth]{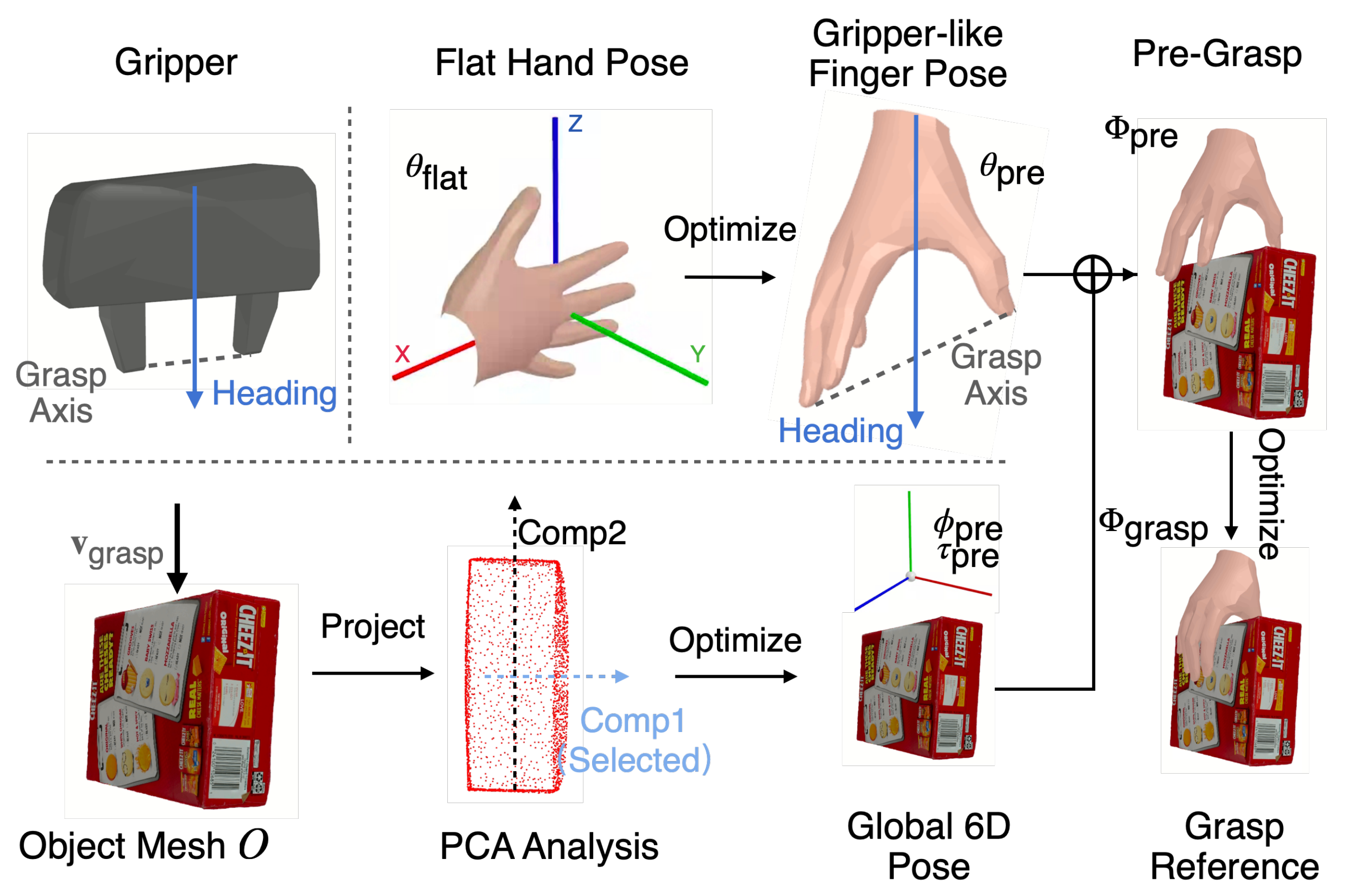}
\end{center}
   \vspace{-0.25cm}
   \caption{\textbf{Grasp Reference Generation.} Given a grasp direction $\mathbf{v}_{\text{grasp}}$ and an object mesh $\bm{O}$, our method generates a pre-grasp pose $\mathbf{\Phi}_{\text{pre}}$ and grasp reference pose $\mathbf{\Phi}_{\text{grasp}}$.}
   \vspace{-1.2em}
\label{fig:gripper}
\end{figure}

\item \textbf{6D Global Pre-Grasp Pose $(\bm{\tau}_{\text{pre}}, \bm{\phi}_{\text{pre}})$.} We visualize the process of determining the global 6D wrist pre-grasp pose in \figref{fig:gripper}. We begin by sampling 3,000 points from the object mesh and identify the sampled point furthest from the object center along the grasp direction. The pre-grasp translation $\bm{\tau}_{\text{pre}}$ is computed as the distance between the object center and the furthest point with an added offset, ensuring there is no collision with the object. To get the pre-grasp global orientation $\bm{\phi}_{\text{pre}}$, we first align the hand's heading with the grasp direction $\mathbf{v}_\text{grasp}$ (cf. \figref{fig:gripper}). We then rotate the hand to grasp the slimmest part of the object. To this end, we project the object points onto a 2D plane orthogonal to the given grasp direction. We run PCA analysis on the projected 2D point set, which yields two principal components. We choose the component with the lower variance that corresponds to the object's narrowest width. Subsequently, the hand is rotated such that the grasp axis aligns with this narrowest segment.

\end{itemize}

\noindent\textit{2) Grasp Reference Pose} $\bm{\Phi}_{\text{grasp}}.$
The second part of our optimization generates a grasp reference pose. It takes as input the hand mesh $\bm{{H}}$, the object mesh $\bm{{O}}$, and the grasp direction $\mathbf{v}_{\text{grasp}}$. We initialize the hand in the pre-grasp pose $\bm{\Phi}_{\text{pre}}$ from the previous stage. Then, multiple losses are optimized to determine the grasp reference pose $\bm{\Phi}_{\text{grasp}}$:
\begin{equation}
\begin{aligned}
\bm{\Phi}_{\text{grasp}} &= \mathop{\argmin}\limits_{\bm{\Phi}} \mathcal{L}(\bm{{H(\bm{\Phi};\bm{\Phi_\text{pre}})}},\bm{{O}}, \mathbf{v}_{\text{grasp}}),\\
\mathcal{L} &= \alpha \mathcal{L}_{\text{DP}} + \beta \mathcal{L}_{\text{C}} + \gamma \mathcal{L}_{\text{DF}} +  \delta \mathcal{L}_{\text{Ctrl}}.
\end{aligned}
\end{equation}
In the following computations, the object and hand vertices are sampled from the mesh surfaces only. The loss function $\mathcal{L}$ comprises the following components:
\begin{itemize}[left=0.0em]
\item \textbf{Dual Penetration Loss $\mathcal{L}_{\text{DP}}$}. We employ a hand-centric penetration loss to minimize penetration during the optimization, similar to \cite{Jiang2021HandObjectCC}. We identifiy the object vertices that are situated inside the hand mesh and compute the penetration loss as the sum of the distances between these vertices and their nearest hand surface vertices:
\begin{equation}
    \begin{aligned}
            \mathcal{L}_{\text{DP}} = \sum_i \mathcal{L}^2(\mathbf{o}_i^{\text{in}}, \mathbf{h}^{\text{closest}}_i) +  \sum_j \mathcal{L}^2(\mathbf{h}_j^{\text{in}}, \mathbf{o}^{\text{closest}}_j),
    \end{aligned}
    \label{eq:dp_loss}
    \vspace{-1mm}
\end{equation}
where $\mathbf{o}_i^{\text{in}} \in \bm{O}$ is the $i$-{th} object point inside the hand mesh and $\mathbf{h}^{\text{closest}}_i \in \bm{H}$ is its closest vertex on the hand surface. While the hand-centric loss mitigates penetration by urging the object vertices toward their closest hand vertices, it is insufficient when entire fingertips are immersed within the object. To address this limitation, we additionally introduce an object-centric penetration loss, implemented in a symmetric manner (second term in \eqnref{eq:dp_loss}).

\item \textbf{Contact Loss $\mathcal{L}_{\text{C}}$}. The contact loss ensures that the hand closely approaches and establishes substantial contact with the object, resulting in a stable grasp. It measures the distance between the hand vertices and their closest corresponding object surface vertices:
\begin{equation}
    \mathcal{L}_{\text{C}} = \sum_k \mathcal{L}^2(\mathbf{h}_k, \mathbf{o}_k^{\text{closest}}),
    \vspace{-3mm}
\end{equation} 
where $\mathbf{h}_k \in \bm{H}$ is the $k$-th hand vertex and $\mathbf{o}_k^{\text{closest}} \in \bm{O}$ is its closest point on the object mesh.
\item \textbf{Dynamic Fingertip Loss $\mathcal{L}_{\text{DF}}$}. This loss mimics the human grasping process. It is calculated based on the distance between the thumb tip and the other four fingertips:
\begin{equation}
    \begin{aligned}
            \mathcal{L}_{\text{DF}} &= k \sum_{l} \mathcal{L}^2(\mathbf{p}_{\text{thumb}}, \mathbf{p}_l).
    \end{aligned}
    \vspace{-3mm}
\end{equation}
$\mathbf{p}_{\text{thumb}}$ is the 3D joint position of the thumb tip and $\mathbf{p}_{l}$ represent 3D joint positions of the other four fingertips. $k$ is the dynamic coefficient, which is negative in the early stages of the optimization (step$<$100) to keep the hand open. In later stages, the coefficient is positive to close the hand towards a stable grasp.

\item \textbf{Control Loss $\mathcal{L}_{\text{Ctrl}}$}. This loss is designed to ensure that the grasp direction will not deviate from the pre-defined direction during the optimization process. We compute it as the cosine similarity between the wrist vector $\mathbf{v}_{\text{wrist}}$, i.e., the vector pointing from the wrist to the object center, and the grasp direction $\mathbf{v}_{\text{grasp}}$: 
\begin{equation}
    \mathcal{L}_{\text{Ctrl}} = 1 - \frac{\mathbf{v}_{\text{wrist}} \cdot \mathbf{v}_{\text{grasp}}}{|\mathbf{v}_{\text{wrist}}| |\mathbf{v}_{\text{grasp}}|}.
\end{equation}

\end{itemize}
The learning rate is initially set to 0.003 and decays by 10\% every 100 steps. The entire optimization process takes 500 steps to obtain the final grasping pose. We set the coefficients $\alpha$, $\beta$, $\gamma$, $\delta$ to 1.5, 3, 0.1, and 1, respectively.

\subsection{Handover Motion Generation}
 To generate handover motions, we pass the grasp reference pose $\bm{\Phi}_{\text{grasp}}$ to our improved variant of D-Grasp \cite{christen2022dgrasp} and initialize the hand in the pre-grasp pose $\bm{\Phi}_{\text{pre}}$. The D-Grasp model takes as input the grasp reference pose and a target 6D object pose. It then generates human motions that approach, grasp, and bring the object into the target pose. In contrast to vanilla D-Grasp, we augment the observation space with information about the object shape to make it more generalizable to unseen objects. Specifically, we compute the signed-distance information \cite{Zhang2021ManipNetNM} by sampling the object's signed-distance field for each hand joint, which we add to D-Grasp by concatenating it to the original observation space \cite{christen2022dgrasp}.

 \begin{table*}[t]
 \centering
 \small
    \aboverulesep=0ex 
   \belowrulesep=0ex 

 \caption{Benchmark Evaluation on HandoverSim and on the Synthetic Test Set.}
 \vspace{-0.1cm}
\resizebox{0.87\linewidth}{!}{%
\begin{tabular}{l|l|l|cccc|cccc}

  \hline
  & & & \multicolumn{4}{c|}{HandoverSim Test Set (DexYCB)} & \multicolumn{4}{c}{Our Synthetic Test Set} \\ \midrule
  & \multirow{2}{*}{Method} & \multirow{2}{*}{Training Data} & \multirow{2}{*}{success (\%)} & \multicolumn{3}{c|}{failure (\%)} & \multirow{2}{*}{success (\%)} & \multicolumn{3}{c}{failure (\%)} \\
  & & & & contact & drop & timeout & & contact & drop & timeout \\
  \hline
  \parbox[t]{2mm}{\multirow{4}{*}{\rotatebox[origin=c]{90}{w/ hold}}}
  & GA-DDPG~\cite{wang:corl2021b} & ShapeNet~\cite{chang2015shapenet} & 50.00         & ~\textbf{4.86} &  19.44        &  25.69 &   31.22   &  16.39      &   43.85    &   ~\textbf{8.54}    \\
  & \gc GA-DDPG finetuned~\cite{christen2023handover} & \gc DexYCB~\cite{chao:cvpr2021} & \gc 57.18        & \gc ~6.48 &  \gc 27.08          &  \gc ~9.26 &   \gc 40.67  &  \gc  15.70    &  \gc 36.24   &   \gc  ~7.38  \\ 
  & Christen et. al~\cite{christen2023handover}  & DexYCB~\cite{chao:cvpr2021} &  \textbf{75.23} & ~9.26 & \textbf{13.43} &  ~\textbf{2.08} & 46.57   & 13.90 &  28.68  &  10.85   \\
  & \gc Christen et. al~\cite{christen2023handover} & \gc Our Synthetic & \gc 71.51  & \gc ~7.87 & \gc 15.30 & \gc ~5.32  & \gc \textbf{56.25} & \gc ~\textbf{8.06} & \gc \textbf{23.50} & \gc 12.17 \\
\hline

  \parbox[t]{2mm}{\multirow{4}{*}{\rotatebox[origin=c]{90}{w/o hold}}}
  
  & GA-DDPG~\cite{wang:corl2021b} & ShapeNet~\cite{chang2015shapenet} &  36.81         &  ~9.03          &  25.00          &  29.17  & 23.90 & 18.26  & 48.67   &  ~9.17 \\
  & \gc GA-DDPG finetuned~\cite{christen2023handover} & \gc DexYCB~\cite{chao:cvpr2021} & \gc 54.86         & \gc ~\textbf{6.71} &  \gc 26.39          & \gc 12.04 & \gc 37.53 & \gc 12.32      & \gc 35.03  & \gc  16.44  \\
  &  Christen et. al~\cite{christen2023handover} & DexYCB~\cite{chao:cvpr2021} &   68.75  & ~8.80 &  17.82 &   ~\textbf{4.63} & 43.20 & 10.92 & 32.84   & 13.02 \\ 
 
  & \gc Christen et. al~\cite{christen2023handover} & \gc Our Synthetic & \gc \textbf{70.60} & \gc ~7.18 & \gc \textbf{16.67} & \gc ~5.56  & \gc \textbf{55.94} & \gc ~\textbf{7.53} & \gc \textbf{25.89} & \gc \textbf{10.63} \\  
\bottomrule
  
\end{tabular}
}
\vspace{-0.25cm}

 \label{tab:quantitative}
\end{table*}

 

We generate a training set of grasp reference poses with our optimization on the DexYCB~\cite{chao:cvpr2021} object set, which we use as guidance to train D-Grasp. After training the model, we generate grasp reference poses on a larger variety of objects. We synthesize human motions by passing these grasp pose references to the trained D-Grasp model. As our optimization allows control of the approaching direction, we sample grasp directions that are pointing towards the robot. Furthermore, we sample random target object 6D poses within the robot's workspace which serve as handover locations. Lastly, we filter out sequences that fail to grasp the object and reach the target 6D object pose.




\section{Augmenting Handover Training}
\label{sec:aug_h2r}
To train the robot, we follow the framework in \cite{christen2023handover}. Instead of training with trajectories from the DexYCB dataset \cite{chao:cvpr2021}, we simulate the humans in the training environment using our synthetic data. The synthetic human motions are replayed in the simulation during training, following the HandoverSim procedure \cite{chao:icra2022}. Our method takes as input egocentric RGB-D images, from which we compute a segmented point cloud (see \figref{fig:method_overview}). If the hand is occluded by the object, we use the last frame where the hand was visible. We then pass the point cloud through PointNet++ \cite{Qi2017PointNetDH} to compute a feature that serves as input to our control policy. The control policy is a neural network that predicts actions that are applied to the robot. Given the updated state, the new point cloud is computed and passed to our policy. The training follows a two-stage procedure. In the pre-training stage, we train in a setting where the human has come to a stop before the robot starts moving. This allows us to leverage expert trajectories from motion and grasp planning \cite{wang:rss2020}, which uses ACRONYM \cite{eppner:acronym2020} to select grasps. To avoid collisions between the robot and the human, we sample grasps that are opposed to the input direction used in the static grasp generation (cf. \secref{sec:static_grasp_gen}). In the fine-tuning stage, we train the robot in a setting where the human and robot move simultaneously. Since we cannot use open-loop motion and grasp planning in this setting, we utilize a frozen version of the pre-trained policy as expert \cite{christen2023handover}. 
Our control policy is trained in actor-critic fashion using a mix of RL-based, behavior cloning, and auxiliary losses as proposed in \cite{wang:corl2021b}. We refer the reader to \cite{christen2023handover} for more details about the overall training procedure and the definition of the losses.

\section{Experiments}
\label{sec:experiments}

\subsection{Experimental Details}
We generate a train and test set of human handover motions using our method on a subset of ShapeNet objects \cite{chang2015shapenet}. We adjust the size of the objects based on the dimensions specified in ACRONYM~\cite{eppner:acronym2020}. To eliminate objects that are too large to grasp for the gripper, we exclude those with a minimal width exceeding 0.15m along the grasp direction.
Our train set comprises 1175 objects and a total of 2230 right-handed handover motions, whereas our test set contains 1174 objects and 4436 handover motions. The test set also includes left-handed motions, which we generate by mirroring the synthesized right-handed motions. As target object 6DoF handover poses, we randomly sample position offsets from the object's initial position within a range of $[-15, 15]$ cm in $x$- and $y$-directions and $[10, 35]$ cm in $z$-direction. For a fair comparison between training on real motion capture and synthetic motions, we use the same training procedure and hyperparameters from \cite{christen2023handover} for both variants. We use a single NVIDIA V100 GPU for training and observe similar convergence times in both data settings.

\subsection{Baselines}
We experiment with two relevant grasping policies~\cite{christen2023handover,wang:corl2021b}. GA-DDPG \cite{wang:corl2021b} is a method for vision-based grasping of rigid objects. Christen \etal~\cite{christen2023handover} is a learning-based method for human-to-robot handovers from point clouds. We use their pre-trained models for evaluation. For \cite{christen2023handover}, we also train on our synthetic data as described in \secref{sec:aug_h2r}. Furthermore, we include the version of GA-DDPG which was trained in the HandoverSim environment following \cite{christen2023handover}. 


\subsection{Metrics} We follow the efficacy metrics in HandoverSim \cite{chao:icra2022}. We report the overall success rate (\textit{success}). A handover is considered a success if the robot grasps the object and moves it to a goal location without dropping or colliding with the human. We distinguish between the three failure cases of human collision (\textit{contact}), object dropping (\textit{drop}), and timeout if the object is not reached (\textit{timeout}). Since we do not focus on improving the efficiency of handovers in this paper, we omit the efficiency metrics from the experiments. 

\subsection{Benchmark Evaluation}
In this experiment, we compare our framework (Christen \etal~\cite{christen2023handover} trained with synthetic data) against baselines on the HandoverSim~\cite{chao:icra2022} test split (i.e., with real human motions from DexYCB~\cite{chao:cvpr2021}). Furthermore, we conduct evaluations on our new synthetic test set to assess generalization to unseen objects and human motions at a larger scale. We report the results in \tabref{tab:quantitative} and indicate the dataset each model was trained on. We differ between the \textit{w/ hold} setup, where the robot only starts moving once the human has stopped, and the \textit{w/o hold} setup, where the robot and the human move at the same time. Please see our supplementary video for qualitative examples of our method and the baselines.

\noindent\textbf{HandoverSim}
Our method outperforms the GA-DDPG \cite{wang:corl2021b} baselines, and reaches comparable performance with Christen \etal~\cite{christen2023handover} trained on DexYCB, e.g., a success rate of 70.60\% for our data and 68.75\% for DexYCB data in the \textit{w/o hold} setting. This result is important, as it shows that using purely synthetic human motion data can match the baseline trained on real human motion data. There is a slight drop in performance for synthetic training data compared to DexYCB in the \textit{w/ hold} setting (from 75.23\% to 71.51\%), which we hypothesize is because the HandoverSim test objects are included in the DexYCB train set, whereas our synthetic data does not contain any of the test objects.

\noindent\textbf{Synthetic Test Set}
We compare against baselines on the new synthetic test set that includes 1174 unseen objects (\tabref{tab:quantitative} right). Notably, the success rate of the baselines drops when evaluated on a large set of unseen objects. In contrast, training on our synthetic data has significantly higher success rates in both the \textit{w/ hold} and the \textit{w/o hold} setting (e.g., a 20\% relative increase in success rate over the most related baseline \cite{christen2023handover}). This indicates that our synthetic training set improves generalization to unseen objects. The decrease in success rate for all methods is expected, as the test set includes unseen objects and hence a much wider variety of different shapes. While the human-robot collisions (\textit{contact}) remain relatively low on the synthetic test set, the object drop rate increases the most, e.g., in the \textit{w/o hold} setting from 17.82\% on HandoverSim to 32.84\% on the synthetic test set for \cite{christen2023handover} trained on DexYCB. This shows that the methods struggle to find feasible grasps on unseen objects. 

\begin{table}[!]
 \vspace{-0mm}
 \centering
 \small
  \caption{\text{Ablation of our framework.}}
   \vspace{-1mm}
  \resizebox{0.85\columnwidth}{!}{%
 \begin{tabular}{l|cccc}

  \hline
  \multirow{2}{*}{Synthetic Test Set} & \multirow{2}{*}{success (\%)} &  \multicolumn{3}{c}{failure (\%)} \\
  & & contact & drop & timeout \\
  \midrule

   w/ GraspTTA \cite{Jiang2021HandObjectCC} & 50.33 & ~\textbf{6.12} & 31.89  &  11.64         \\
   \gc Ours - 25\% & \gc 47.23  & \gc ~8.08 & \gc 25.75 & \gc 18.93         \\

   Ours - 50\%  & 49.21  & 10.75 & 27.22 & 12.80          \\

  \gc Ours & \gc \textbf{55.94} & \gc ~7.53 & \gc 25.89 & \textbf{10.63} \gc \\ 
  \hline
 \end{tabular}
 }

 \vspace{-2mm}
 \label{tab:ablation}
\end{table}
\subsection{Ablations}
We ablate our synthetic data generation pipeline by comparing it against a variant where we use GraspTTA \cite{Jiang2021HandObjectCC} (\textit{w/ GraspTTA)} instead of our method to generate grasp references for D-Grasp. Furthermore, we analyze the influence of the size of the synthetic train set on the test performance. We report the results on the synthetic test set in \tabref{tab:ablation}. We find that a larger training set of synthetic data helps with generalization to unseen objects and human motions, as shown by the decreased performance when only 50\% or 25\% of the synthetic training set are used for training. Our grasp generator can generate more suitable grasps for handovers than GraspTTA, as shown by the relative increase of 10\% in success rate. This implies that the conditioning on the grasping direction is favorable for handover policy training. Note that the ShapeNet objects used are within the training distribution of GraspTTA, as it is trained on Obman \cite{hasson19obman}, and it is likely to perform worse on unseen objects.

\subsection{Sim-to-Real Evaluation}
Finally, we transfer the policy trained with our synthetic dataset onto a real robotic platform (system \textit{A}) and compare it with the policy trained on DexYCB from~\cite{christen2023handover} (system \textit{B}). We seek to answer the question: \textit{Can a person differentiate these two systems from interacting with them?} To answer this, we run a human evaluation with 8 participants. For each participant, the experiment consists of two phases. In the first phase, we let the participant hand over three YCB objects, each to both systems once. After each handover, we inform the participant which system is used (\textit{A} or \textit{B}). In the second phase, we use the 10 household objects selected in~\cite{yang:icra2021} (see Fig.~6 in~\cite{yang:icra2021}) and ask each participant to hand over each object to the robot just once. We randomly sample a system for each object and equally distribute the choices of the two systems (i.e., \textit{A} for 5 objects and \textit{B} for the other 5). In this phase, we do not disclose the chosen system to the participant. After each handover, we ask the participant to make a guess of the chosen system based on the interaction. In the end, we found the two systems both performing competently, exhibiting over 85\% handover success rate (45/50 for \textit{A} and 48/50 for \textit{B}). The classification accuracies from the participants are: (8/10, 6/10, 4/10, 10/10, 10/10, 6/10, 6/10, 10/10) (random guessing is expected to get 5/10). Four of them have an accuracy less or equal to 6/10, struggling to tell apart the two systems. Among the other four, two of them answered ``felt equal'' in a forced choice question between ``preferred \textit{A}'', ``preferred \textit{B}'', and ``felt equal''. They also commented that the systems can be distinguished due to their subtly different tendencies in the approaching direction (the baseline policy tends to have a slight left tilt before handover). This may have resulted from the randomness in training. Overall, this result suggests that our system trained purely on synthetic data is performing closely to a system trained on real data.

\section{Conclusion}
We have introduced a framework to generate synthetic human motions for handover training. Our method combines a non-learning based grasp optimizer with an RL-based policy. We have generated a synthetic training set and demonstrated that training with our generated motions reaches a similar performance to training with motion capture data, both in simulation and on a real system. Moreover, we have shown that training with our synthetic data generalizes better to unseen objects on a large-scale synthetic test set. While this work focuses on generalizability to new objects, future work can investigate the generalizability to different human motions or robot morphologies. For example, it is interesting to explore longer interactions \cite{wang2024genh2r}, the integration of full-body synthetic humans \cite{braun2023physically}, or more challenging human-robot interactions such as two-handed handovers and articulated objects \cite{zhang2024artigrasp}.

\bibliographystyle{IEEEtran}

\bibliography{IEEEabrv,references}

\end{document}